# Tối ưu hiệu suất tốc độ Động cơ Servo DC sử dụng bộ điều khiển PID kết hợp mạng Nơ-ron

Lê Tiểu Niên[1,2], Phạm Văn Cường[1], Nguyễn Phúc Ánh[1] & Vũ Ngọc Sơn[1,*]

[1] Khoa điện, Đại học Công nghiệp Hà Nội, Hà Nội, Việt Nam
[2] Khoa điện – điện tử, Đại học Công nghệ Đông Á, Bắc Ninh, Việt Nam
Email: vungocson.haui.271199@gmail.com

*Abstract* — Động cơ DC đã được sử dụng phổ biến trong nhiều ứng dụng công nghiệp, từ các robot khớp nối nhỏ với nhiều bậc tự do, đến các thiết bị điện gia dụng và phương tiện di chuyển như xe điện và tàu hỏa. Chức năng chính của những động cơ này là đảm bảo hiệu suất định vị và tốc độ ổn định cho các hệ thống cơ khí, dựa trên các phương pháp điều khiển đã được thiết kế trước đó. Tuy nhiên, việc đạt được hiệu suất tốc độ tối ưu cho động cơ servo gặp nhiều khó khăn do tác động của tải trọng bên trong và bên ngoài, làm ảnh hưởng đến sự ổn định của đầu ra. Để tối ưu hóa hiệu suất tốc độ của động cơ Servo DC, phương pháp điều khiển kết hợp giữa bộ điều khiển PID và mạng nơ-ron nhân tạo đã được đề xuất. Bộ điều khiển PID truyền thống có ưu điểm ở cấu trúc đơn giản và khả năng điều khiển hiệu quả trong nhiều hệ thống, nhưng lại gặp khó khăn khi đối mặt với các thay đổi phi tuyến tính và không xác định. Mạng nơ-ron được tích hợp để điều chỉnh các tham số của bộ PID theo thời gian thực, giúp hệ thống thích ứng với các điều kiện hoạt động khác nhau. Kết quả mô phỏng và thực nghiệm đã chứng minh rằng phương pháp đề xuất không chỉ cải thiện đáng kể khả năng theo dõi tốc độ và tính ổn định của động cơ, mà còn đảm bảo đáp ứng nhanh, sai số xác lập về 0 và loại bỏ hiện tượng vọt lố. Phương pháp này mang lại tiềm năng ứng dụng cao trong các hệ thống điều khiển động cơ servo yêu cầu độ chính xác và hiệu suất cao.

## I. GIỚI THIỆU

Động cơ DC đóng vai trò quan trọng trong các hệ thống điều khiển tự động, đặc biệt trong các ứng dụng công nghiệp và dân dụng. Những ứng dụng này bao gồm robot công nghiệp, máy CNC, thiết bị gia dụng và các phương tiện di chuyển như xe điện và tàu hỏa. Khả năng điều khiển chính xác về vị trí và tốc độ của động cơ Servo DC khiến nó trở thành lựa chọn lý tưởng cho các hệ thống yêu cầu độ tin cậy và hiệu suất cao. Tuy nhiên, việc duy trì tốc độ ổn định và chính xác của động cơ Servo DC đối mặt với nhiều thách thức, nhất là khi có tác động của các yếu tố phi tuyến và không xác định, chẳng hạn như biến đổi tải trọng hoặc điều kiện môi trường thay đổi [1,2].

Bộ điều khiển PID truyền thống được sử dụng rộng rãi trong các hệ thống điều khiển tuyến tính nhờ cấu trúc đơn giản và hiệu quả cao. PID hoạt động dựa trên việc điều chỉnh ba tham số: tỷ lệ (P), tích phân (I) và vi phân (D), giúp giảm thiểu sai số và nhanh chóng đưa hệ thống vào trạng thái ổn định. Tuy nhiên, việc áp dụng bộ điều khiển PID đòi hỏi phải hiểu sâu sắc về hệ thống và các thông số điều khiển, cũng như cần phải tinh chỉnh các thông số để đạt được hiệu suất tối ưu. Quá trình này bao gồm việc xác định các giá trị thích hợp cho các mức tăng tỷ lệ, tích phân và đạo hàm, cùng với việc thử nghiệm và điều chỉnh chúng trong quá trình vận hành thực tế [3-5]. Điều này không chỉ đòi hỏi chuyên môn kỹ thuật mà còn đòi hỏi kinh nghiệm thực tế để thực hiện các điều chỉnh hiệu quả.

Trong nghiên cứu này, mạng nơ-ron nhân tạo được áp dụng để tối ưu hóa bộ điều khiển PID nhằm điều khiển tốc độ động cơ Servo DC. Hệ thống điều khiển sử dụng mô hình CE110 Servo Trainer, một hệ thống nghiên cứu và đào tạo phổ biến trong lĩnh vực điều khiển tự động, được trang bị các cảm biến và cơ cấu chấp hành chính xác, cho phép đánh giá và thử nghiệm các thuật toán điều khiển trong điều kiện thực tế [11].

Mục tiêu của nghiên cứu là chứng minh hiệu quả của việc kết hợp giữa PID và mạng nơ-ron nhân tạo trong việc cải thiện hiệu suất điều khiển. Cụ thể, nghiên cứu tập trung vào việc giảm thời gian ổn định và sai số trạng thái bền so với bộ điều khiển PID truyền thống (Kuhn) và một bộ điều khiển được chọn ngẫu nhiên bởi người điều khiển. Sự kết hợp này được kỳ vọng sẽ mang lại khả năng điều chỉnh linh hoạt và chính xác hơn, giúp hệ thống nhanh chóng đạt được trạng thái ổn định và duy trì hiệu suất cao trong suốt quá trình vận hành. Ngoài ra, nghiên cứu cũng xem xét các yếu tố ảnh hưởng đến quá trình tối ưu hóa, bao gồm các tham số mạng nơ-ron như số lượng lớp ẩn, hàm kích hoạt, và tốc độ học, nhằm tìm ra cấu hình tối ưu cho bộ điều khiển. Bên cạnh đó, các điều kiện hoạt động khác nhau của hệ thống, chẳng hạn như tải trọng và tốc độ yêu cầu, cũng được đánh giá. Kết quả của nghiên cứu không chỉ cung cấp những hiểu biết sâu sắc về tính khả thi của việc ứng dụng mạng nơ-ron để tối ưu hóa bộ điều khiển PID mà còn mở ra những cơ hội phát triển các hệ thống điều khiển tự động tiên tiến. Những phát hiện từ nghiên cứu này được kỳ vọng sẽ góp phần nâng cao hiệu suất và độ tin cậy của các hệ thống điều khiển trong ứng dụng công nghiệp và nghiên cứu khoa học.

Phương pháp điều khiển kết hợp giữa PID và mạng nơ-ron nhân tạo được đề xuất nhằm mục tiêu tối ưu hóa hiệu suất điều khiển tốc độ của động cơ Servo DC trong các điều kiện hoạt động đa dạng. Các kết quả mô phỏng và thực nghiệm sẽ minh chứng cho hiệu quả của phương pháp này trong việc giảm thời gian ổn định và sai số ở trạng thái ổn định so với các bộ điều khiển PID truyền thống.

Cấu trúc của bài báo gồm các phần sau: Phần 2 cung cấp tổng quan về động cơ Servo DC và phương pháp điều khiển PID kết hợp mạng nơ-ron nhân tạo. Phần 3 mô tả mô hình thực nghiệm và quy trình áp dụng. Phần 4 trình bày kết quả mô phỏng và thực nghiệm. Cuối cùng, phần 5 tổng kết và đề xuất hướng phát triển trong tương lai.

## II. MẠNG PID NƠ-RON

Mạng nơ-ron PID là một hệ thống bao gồm các nơ-ron với ba thành phần chính: nơ-ron P, nơ-ron I, và nơ-ron D, tương ứng với các chức năng tỉ lệ, tích phân và vi phân. Nơ-ron P thực hiện việc điều chỉnh dựa trên sai số hiện tại, nơ-ron I xử lý sai số tích lũy theo thời gian, và nơ-ron D dự đoán xu hướng thay đổi của sai số. Hệ thống này có thể được sử dụng để tạo ra một bộ điều khiển PID thay thế cho bộ điều khiển PID truyền thống trong các ứng dụng điều khiển sản xuất. Mạng nơ-ron PID tích hợp thuật toán điều khiển của PID với khả năng tự học của mạng nơ-ron, giúp bộ điều khiển có khả năng điều chỉnh thông minh hơn. Cấu trúc của mạng nơ-ron PID bao gồm ba lớp với cấu hình 3-3-1, đảm bảo khả năng nhận các giá trị đầu vào và phản hồi, đồng thời xử lý thông tin qua các tầng nơ-ron để điều chỉnh hệ thống.

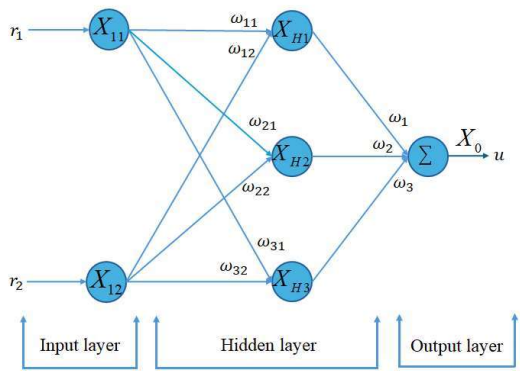

Hình 1. Mạng Nơ-ron PID

Từ hình 1 có thể thấy rằng mạng nơ-ron PID là một mạng nơ-ron truyền thẳng, bao gồm 2 tín hiệu đầu vào lần lượt là tín hiệu tốc độ mong muốn và tín hiệu phản hồi tốc độ, và cũng là các nơ-ron $X_{11}$ và $X_{12}$ ở lớp đầu vào, ba nơ-ron $X_{H1}$, $X_{H1}$ và $X_{H3}$ ở lớp ẩn, và một nơ-ron $X_0$ ở lớp đầu ra. Lớp đầu vào có ba nút đầu vào, và nhiệm vụ chính của nó là chuyển tín hiệu đầu vào đến các nơ-ron trong lớp ẩn. Trong số ba nơ-ron ở lớp ẩn, $X_{H1}$ đóng vai trò là nơ-ron P, $X_{H2}$ là nơ-ron I, và $X_{H3}$ là nơ-ron D. Nơ-ron của lớp đầu ra là một nơ-ron tiêu chuẩn của mạng nơ-ron truyền thẳng. Chức năng của mạng nơ-ron PID được mô tả như sau. Để xem xét trạng thái đầu vào và đầu ra của mạng nơ-ron PID tại thời điểm lấy mẫu k:

\* Lớp đầu vào

$$X_{1i}(k) = r_i(k) \tag{1}$$

Trong đó:

$r_i(k)$, với i=1,2, là các tín hiệu đầu vào tại thời điểm lấy mẫu k; lần lượt là tín hiệu tốc độ mong muốn và tín hiệu phản hồi tốc độ.

\* Lớp ẩn

Lớp ẩn là lớp quan trọng trong mạng nơ-ron PID, nơi thực hiện các phép toán PID. Trong lớp ẩn, $X_{H1}$ là nơ-ron P (tỉ lệ), $X_{H2}$ là nơ-ron I (tích phân) và $X_{H3}$ là nơ-ron D (vi phân). Chức năng của các nơ-ron này sẽ được mô tả chi tiết như sau: (1) Nơ-ron tỉ lệ

Hàm đầu vào và hàm đầu ra của nơ-ron tỉ lệ $X_{H1}$ có thể được biểu diễn qua các công thức (2) (3) sau:

$$u_{H1}(k) = u_P(k) = \sum_{i=1}^{2} \omega_{1i} X_{1i}(k) \tag{2}$$

Trong đó, $\omega_{1i}$, với i=1,2 lần lượt là các giá trị trọng số giữa $X_{H1}$ và lớp đầu vào.

$$X_{H1}(k) = \begin{cases} 1 & , u_{H1}(k) > 1 \\ u_{H1}(k) & , -1 < u_{H1}(k) < 1 \\ -1 & , u_{H1}(k) < -1 \end{cases} \tag{3}$$

(2) Nơ-ron tích phân

Hàm đầu vào và hàm đầu ra của nơ-ron tích phân $X_{H2}$ có thể được biểu diễn qua các công thức (4), (5).

$$u_{H2}(k) = u_I(k) = \sum_{i=1}^{2} \omega_{2i} X_{1i}(k) \tag{4}$$

Trong đó, $\omega_{2i}$, với i=1,2 lần lượt là các giá trị trọng số giữa $X_{H2}$ và lớp đầu vào.

$$X_{H2}(k) = \begin{cases} 1 & , U_{H2}(k) > 1 \\ u_{H2}(k) + u_{H2}(k-1) & , -1 \leq U_{H2}(k) \leq 1 \\ -1 & , U_{H2}(k) < -1 \end{cases} \tag{5}$$

(3) Nơ-ron vi phân

Hàm đầu vào và hàm đầu ra của nơ-ron vi phân $X_{H3}$ có thể được biểu diễn bằng các công thức sau.

$$u_{H3}(k) = u_D(t) = \sum_{i=1}^{2} \omega_{3i} X_{1i}(k) \tag{6}$$

Trong đó, $\omega_{3i}$, với i=1,2,2 lần lượt là các giá trị trọng số giữa $X_{H3}$ và lớp đầu vào.

$$X_{H3}(k) = \begin{cases} 1 & , U_{H3}(k) > 1 \\ u_{H3}(k) - u_{H3}(k-1) & , -1 \leq U_{H3}(k) \leq 1 \\ -1 & , U_{H3}(k) < -1 \end{cases} \tag{7}$$

\* Lớp đầu ra

Nơ-ron $X_0$ là nơ-ron duy nhất của lớp đầu ra, có chức năng hội tụ ba kết quả từ các nơ-ron P, I, D và sau đó xuất ra tín hiệu điều khiển tổng hợp. Hàm đầu vào và hàm đầu ra của nó có thể được biểu diễn bằng các công thức (8) và (9).

$$u(k) = \sum_{i=1}^{3} \omega_i X_{Hi}(k) \quad (8)$$

Trong đó, $\omega_i$, với i=1,2,3 là các giá trị trọng số giữa lớp đầu ra và lớp ẩn.

$$u(k) = u_{PID}(k) = \omega_1 u_P(k) + \omega_2 u_I(k) + \omega_3 u_D(k) \quad (9)$$

## III. MÔ TẢ HỆ THỐNG

### 3.1. Giới thiệu mô hình

Mô hình thực nghiệm là hệ thống điều khiển động cơ Servo từ Tecquipment [11], bao gồm một động cơ điện, một tải tuyến tính, thiết bị đo và hiển thị tốc độ động cơ, và một tải biến thiên được điều khiển bởi tín hiệu điện.

Mô hình của hệ thống được biểu diễn qua sơ đồ khối như hình 2.

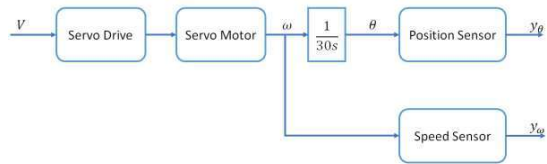

Hình 2. Sơ đồ khối hệ thống điều khiển phản hồi CE110 Servo Trainer

Trong hệ thống này, $V$ là điện áp cấp cho động cơ, $\omega$ là tốc độ trục động cơ, và $\theta$ là vị trí góc của trục động cơ. Các tín hiệu đầu ra $y_\omega$ và $y_\theta$ đại diện cho tín hiệu phản hồi từ cảm biến tốc độ và vị trí, tương ứng [12].

### 3.2. Triển khai trên mô hình thực tế

Mô hình CE110 từ TecQuipment được kết nối với máy tính thông qua thẻ PCIe-6321, cho phép giao tiếp và điều khiển hệ thống hiệu quả. Tốc độ động cơ được điều chỉnh dựa trên tín hiệu điện áp đầu vào, trong khi bộ mã hóa hiển thị tốc độ động cơ hiện tại. Bộ mã hóa cũng truyền đạt các đặc tính quá độ của tốc độ động cơ servo tới máy tính để phân tích và giám sát.

Mô hình hệ thống thực nghiệm CE110 trong nghiên cứu này được thể hiện trong hình 3 gồm:

+ 1 hệ thống CE110 gồm các cổng tín hiệu Encoder, tín hiệu điều khiển cho động cơ Input Control, tải gắn trên động cơ

+ 1 bộ điều khiển PCIe-6321 và các dây nối

+ Hệ thống PC có cổng kết nối với bộ điều khiển PCIe-6321 và phần mềm Matlab/Simulink

Hệ thống được mô phỏng và điều khiển thông qua Matlab/Simulink, với chu kỳ lấy mẫu được thiết lập là $T_s = 10ms$, đảm bảo việc thu thập và phản hồi dữ liệu nhanh chóng. Trong Matlab/Simulink, khối khuếch đại được sử dụng như một khối tỷ lệ để chuyển đổi tín hiệu điện áp đầu ra thành tốc độ động cơ. Hệ số khuếch đại cảm biến được định nghĩa là 1V tương đương với 200 RPM, đảm bảo độ chính xác trong việc điều khiển và giám sát tốc độ động cơ.

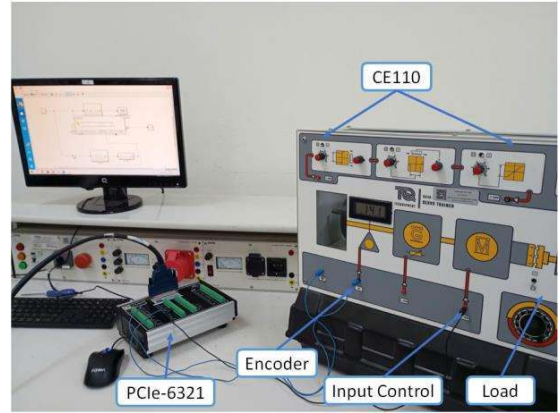

Hình 3. Mô hình CE110 Servo Trainer với bộ điều khiển PCIe-6321

## IV. KẾT QUẢ

### 4.1. Kết quả mô phỏng

Đoạn này trình bày kết quả mô phỏng của hệ thống điều khiển được đề xuất nhằm đánh giá hiệu suất của bộ điều khiển PID Nơ-ron. Việc mô phỏng được thực hiện bằng cách sử dụng Matlab/Simulink dựa trên mô hình lý thuyết đã được mô tả trong [12].

Hàm truyền đạt đại diện cho mô hình là: $G(s) = \frac{0,946}{1+0,4425s} e^{-0,0325s}$

Kết quả phản hồi tốc độ động cơ đã được mô phỏng cho các trường hợp sau:

- Tốc độ động cơ mong muốn là 200 RPM
- Tốc độ động cơ thay đổi theo thời gian
- Tốc độ động cơ dưới các điều kiện tải khác nhau

Các chỉ số hiệu suất của hệ thống điều khiển PID Nơ-ron được so sánh với các chỉ số của bộ điều khiển PID được tính toán dựa trên lý thuyết tổng hợp Kuhn và một bộ điều khiển PID được chọn ngẫu nhiên, như được trình bày trong Bảng 1.

**Bảng 1.** *Tham số PID kinh điển*

| Thông số | Kp | Ki | Kd |
|---|---|---|---|
| PID - Kuhn | 1.057 | 3.125 | 0.08016 |
| PID | 1 | 1 | 1 |

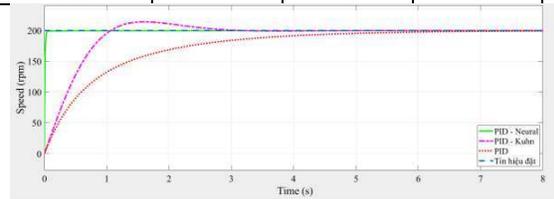

Hình 4. Phản hồi tốc độ động cơ với tốc độ mong muốn 200 RPM

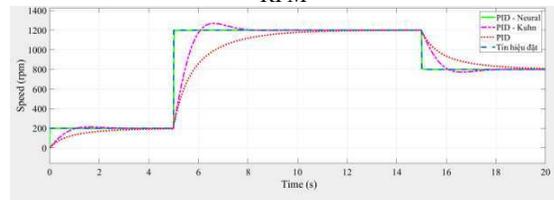

Hình 5. Phản hồi tốc độ động cơ với tốc độ động cơ thay đổi

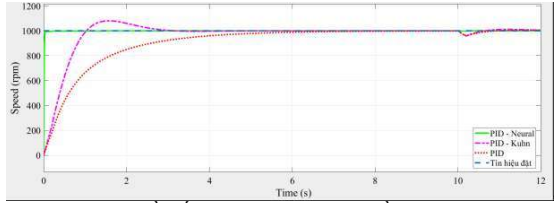

Hình 6. Phản hồi tốc độ động cơ trong điều kiện hoạt động ổn định và tải thay đổi tại t = 10 giây

*Bảng 2. Các tham số chất lượng của mô phỏng*

| Bộ điều khiển | | Độ quá điều chỉnh (%) | Thời gian xác lập (s) | Sai số xác lập |
|---|---|---|---|---|
| PID Nơ-ron | Tốc độ mong muốn 200 vòng/phút | 0 | 0.03 | 0 |
| PID – Kuhn | | 7.15 | 2.19 | 0 |
| PID | | 0 | 4.23 | 0 |
| PID Nơ-ron | Tăng tốc | 0 | 0.03 | 0 |
| PID – Kuhn | | 6.8 | 2.56 | 0 |
| PID | | 0 | 4.78 | 0 |
| PID Nơ-ron | Giảm tốc | 0 | 0.02 | 0 |
| PID – Kuhn | | 6.82 | 1.97 | 0 |
| PID | | 0 | 2.87 | 0 |

Kết quả trong Bảng 3 cho thấy bộ điều khiển PID Nơ-ron mang lại phản hồi tốc độ tối ưu trong tất cả các trường hợp: không có hiện tượng quá điều chỉnh và thời gian ổn định ngắn nhất. Điều này chứng minh rằng PID Nơ-ron không chỉ cải thiện chất lượng phản hồi của hệ thống mà còn vượt trội hơn các bộ điều khiển PID khác, chẳng hạn như PID-Kuhn và PID tiêu chuẩn. Trong các quá trình tăng tốc, giảm tốc và duy trì tốc độ ổn định mong muốn, bộ điều khiển PID Nơ-ron liên tục giữ được mức độ quá điều chỉnh bằng không, thời gian ổn định rất ngắn và không có sai số trạng thái ổn định, cho thấy rõ hiệu quả và hiệu suất tối ưu của nó trong việc điều khiển hệ thống.

**4.2. Kết quả thực nghiệm**

Phần này trình bày kết quả thí nghiệm của bộ điều khiển PID Nơ-ron được áp dụng cho cả mô hình lý thuyết đã thảo luận trong phần kết quả mô phỏng và các thí nghiệm thực tiễn trên hệ thống CE110 Servo Trainer. Sơ đồ hệ thống thực nghiệm được thiết lập và được xây dựng trên Matlab/Simulink nhằm đánh giá hiệu suất của bộ điều khiển PID Nơ-ron trong các tình huống như được thể hiện trong Hình 7.

Trong Hình 7, là sơ đồ hệ thống thực nghiệm và mô phỏng trên Matlab/Simulink với:

+ Hệ thống 1 (phía trên) là hệ thống thực nghiệm trên mô hình CE110 Servo Trainer với các tín hiệu được thu thập từ bộ điều khiển PCIe-6321 như Analog Input, Analog Output và bộ điều khiển PID nơ-ron

+ Hệ thống 2 (phía dưới) là hệ thống mô phỏng với hàm truyền đối tượng của mô hình CE110 Servo Trainer được xây dựng theo công trình [12] và tài liệu hướng dẫn của hãng [11].

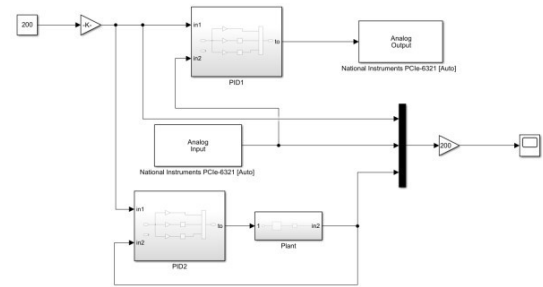

Hình 7. Cấu trúc của hệ thống điều khiển tốc độ động cơ qua PCIe-6321 trên Matlab/Simulink

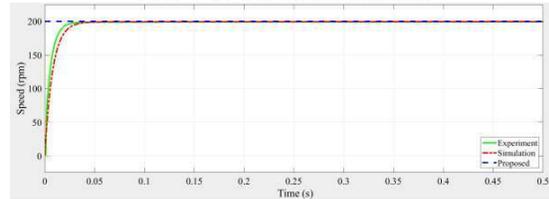

Hình 8. Phản hồi tốc độ động cơ với tốc độ mong muốn 200 RPM

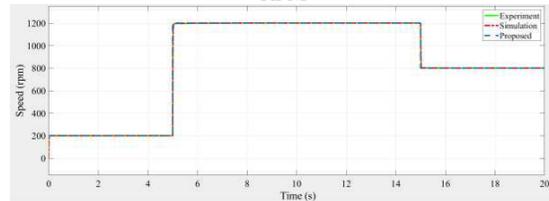

Hình 9. Phản hồi tốc độ động cơ với tốc độ động cơ thay đổi

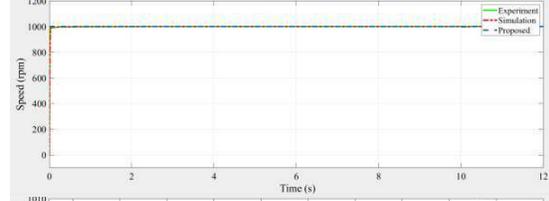

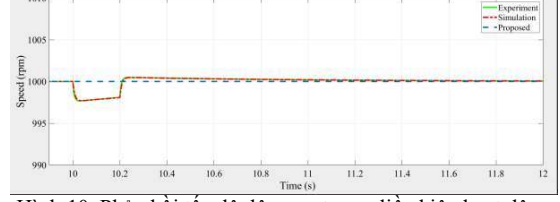

Hình 10. Phản hồi tốc độ động cơ trong điều kiện hoạt động ổn định và tải thay đổi tại t = 10 giây

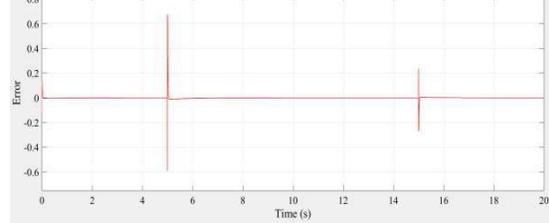

Hình 11. Sai lệch phản hồi tốc độ ở trường hợp thay đổi tốc độ

Kết quả thu được từ cả mô hình lý thuyết và thí nghiệm thực tế cho thấy sự tương đồng cao, chứng minh rằng bộ điều khiển PID Nơ-ron có hiệu quả trong việc điều khiển cả mô hình mô phỏng và hệ thống thí nghiệm. Sự tương thích này cho thấy bộ điều khiển đề xuất đã điều khiển tốc độ mong muốn của hệ thống động cơ servo một cách hiệu quả và giảm thiểu sai số trong cả hai trường hợp. Những kết quả nhất quán giữa mô hình lý thuyết và thí nghiệm thực tế xác nhận tính khả thi và hiệu quả của bộ điều khiển PID Nơ-ron trong việc điều khiển hệ thống động cơ servo trong cả môi trường mô hình và thực tế.

## V. KẾT LUẬN

Nghiên cứu đã đề xuất một phương pháp tối ưu hóa hiệu suất tốc độ cho động cơ Servo DC bằng cách kết hợp bộ điều khiển PID với mạng nơ-ron nhân tạo. Kết quả từ mô phỏng và thực nghiệm cho thấy sự vượt trội của bộ điều khiển PID Nơ-ron so với các phương pháp điều khiển truyền thống như PID-Kuhn và PID tiêu chuẩn dựa trên kinh nghiệm hiểu biết về hệ thống cần điều khiển. Cụ thể, bộ điều khiển PID Nơ-ron không chỉ loại bỏ hiện tượng quá điều chỉnh mà còn đảm bảo thời gian ổn định ngắn và sai số trạng thái ổn định bằng không trong các trường hợp được thử nghiệm như tốc độ mong muốn là hằng số, thay đổi tốc độ, thay đổi tải trong quá trình vận hành.

Sự tương đồng giữa kết quả mô phỏng và thực nghiệm chứng minh tính khả thi và hiệu quả của phương pháp đề xuất trong việc điều khiển động cơ servo, từ đó mở ra hướng nghiên cứu mới cho các ứng dụng trong lĩnh vực điều khiển tự động hóa. Các kết quả này khẳng định tiềm năng của việc áp dụng các công nghệ học máy trong các hệ thống điều khiển phức tạp, đặc biệt là trong các ứng dụng yêu cầu độ chính xác và hiệu suất cao.